\documentclass[runningheads]{llncs}

\usepackage{eccv}

\usepackage{eccvabbrv}

\usepackage{graphicx}
\usepackage{epsfig} 
\usepackage{amsmath} 
\usepackage{amssymb}  
\usepackage{booktabs}
\usepackage{caption}
\usepackage{multirow}
\usepackage{multicol}
\usepackage{wrapfig}
\usepackage{color}
\usepackage{makecell}
\usepackage{marvosym}
\definecolor{lowerred}{RGB}{255,110,180}

\definecolor{lowred}{RGB}{238,18,137}

\newcommand{\ddplus}[1]{\fontsize{0.5pt}{0.05em}\selectfont (\textbf{\textcolor{lowerred}{#1}})}
\newcommand{\dplus}[1]{\fontsize{6pt}{0.1em}\selectfont (\textbf{\textcolor{lowred}{#1}})}

\newcommand{\tabincell}[2]{\begin{tabular}{@{}#1@{}}#2\end{tabular}}

\usepackage[accsupp]{axessibility}  

\usepackage{hyperref}

\usepackage{orcidlink}

\begin{document}

\title{Make Your ViT-based Multi-view 3D Detectors Faster via Token Compression}

\titlerunning{ToC3D}

\author{Dingyuan Zhang\inst{1\textbf{*}}\orcidlink{0009-0001-5022-8172} \and
Dingkang Liang\inst{1\textbf{*}}\orcidlink{0000-0003-3035-1373} \and
Zichang Tan\inst{2}\orcidlink{0000-0002-8501-4123
} \and
\\Xiaoqing Ye\inst{2}\orcidlink{0000-0003-3268-880X} \and
Cheng Zhang\inst{1}\orcidlink{0000-0001-6831-5103} \and
Jingdong Wang\inst{2}\orcidlink{0000-0002-4888-4445
} \and
Xiang Bai\inst{1\textsuperscript{\Letter}}\orcidlink{0000-0002-3449-5940
}
}

\authorrunning{D. Zhang et al.}

\institute{Huazhong University of Science and Technology, Wuhan, China\\
\email{\{dyzhang233, dkliang, xbai\}@hust.edu.cn}\and
Baidu Inc., Beijing, China
}
{\let\thefootnote\relax\footnotetext{\textbf{*} Dingyuan Zhang and Dingkang Liang contributed equally. $\textsuperscript{\Letter}$ Corresponding author.}}
\maketitle

\begin{abstract}

Slow inference speed is one of the most crucial concerns for deploying multi-view 3D detectors to tasks with high real-time requirements like autonomous driving. Although many sparse query-based methods have already attempted to improve the efficiency of 3D detectors, they neglect to consider the backbone, especially when using Vision Transformers (ViT) for better performance. To tackle this problem, we explore the efficient ViT backbones for multi-view 3D detection via token compression and propose a simple yet effective method called \textbf{To}ken\textbf{C}ompression\textbf{3D} (ToC3D). By leveraging history object queries as foreground priors of high quality, modeling 3D motion information in them, and interacting them with image tokens through the attention mechanism, ToC3D can effectively determine the magnitude of information densities of image tokens and segment the salient foreground tokens. With the introduced dynamic router design, ToC3D can weigh more computing resources to important foreground tokens while compressing the information loss, leading to a more efficient ViT-based multi-view 3D detector. Extensive results on the large-scale nuScenes dataset show that our method can nearly maintain the performance of recent SOTA with up to 30\% inference speedup, and the improvements are consistent after scaling up the ViT and input resolution. The code will be made at \url{https://github.com/DYZhang09/ToC3D}.

\end{abstract}

\begin{figure}[t]
\begin{center}
\includegraphics[width=0.95\linewidth]{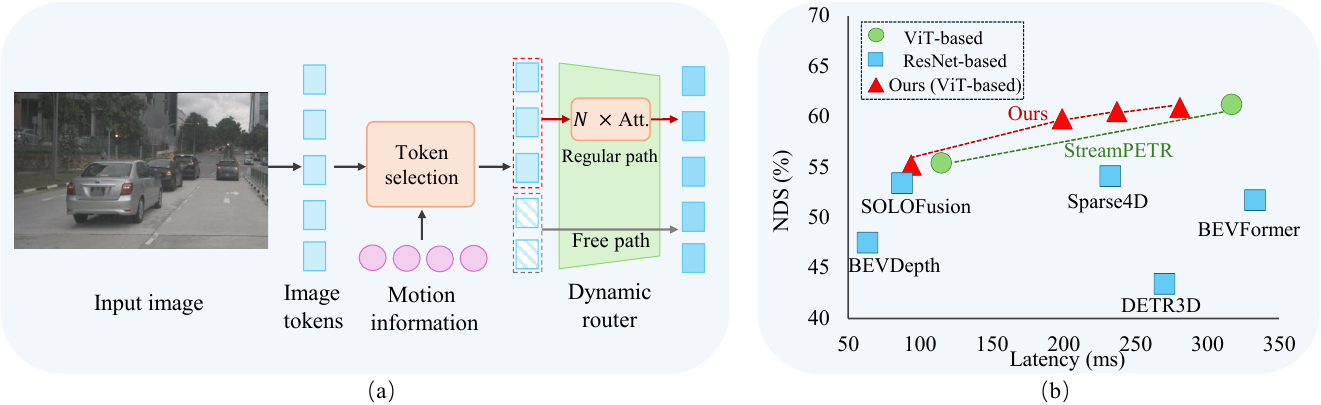}
\end{center}
\caption{
(a) We trim ViTs by focusing on the foreground tokens with the aid of motion cues. (b) Our method reports an ideal trade-off between performance and latency.
}
\label{fig:intro}
\end{figure}

\section{Introduction}
\label{sec:intro}

Multi-view 3D object detection is one of the most fundamental 3D vision tasks crucial for many real-world applications (\eg, autonomous driving~\cite{ma2023camera_od_for_av}), which has gained more effort and achieved great success in recent years.

Existing multi-view 3D object detection methods can be mainly categorized into two types: dense Bird's Eye View (BEV)-based~\cite{huang2022bevpoolv2, huang2022bevdet4d, li2022bevformer} and sparse query-based~\cite{wang2022detr3d, wang2023streampetr, lin2022sparse4d}. The former extracts dense BEV features from images and then interacts with object queries to finish detection, while the latter directly uses sparse object queries to interact with image features, skipping the dense BEV feature extraction. Because sparse query-based methods mainly acquire features of 3D objects instead of whole large-scale scenes, they leverage the sparsity better, and tremendously relax the computing and memory resources requirements. However, this design puts forward higher requirements for image feature quality since sparse query-based methods cannot refine the features in BEV space. Thus, image backbones with better capacity would be beneficial. 

Recently, Vision Transformers (ViTs)~\cite{dosovitskiy2020vit, he2022mae, fang2023eva02} have dominated the vision tasks due to their high capacity, scalability, and flexibility to integrate with multi-modal foundation models. 
For the sake of the performance and flexibility of 3D detection, many sparse query-based multi-view 3D detectors have been trained with advancing pre-trained ViTs. The usage of ViTs has become a trend that is increasingly prevailing. Nowadays, the sparse query-based multi-view 3D detectors~\cite{wang2023streampetr, lin2022sparse4d, liu2023sparsebev} with powerful ViTs have achieved state-of-the-art performance and nearly dominated the leaderboard.

Nevertheless, although sparse query-based methods improve the efficiency by concentrating mainly on foreground objects in the 3D decoder, we experimentally found that the inference speed is not mainly hindered by the 3D decoder but by the ViT backbone. One reason is that existing sparse query-based methods use the ViTs without adjustments, treating foreground 3D objects and background things all the same. Despite the simplicity, we argue that the naive usage of ViT backbones does not obey their design principles: foreground proposals are more significant than background for 3D object detection, and we do not need to model background things in detail. This negligence brings an unnecessary burden, which motivates us to ``trim'' the ViT backbone to achieve better efficiency. 

A simple way is to accelerate the ViT backbones for multi-view 3D detectors via token compression~\cite{rao2021dynamicvit, meng2022adavit, xu2022evovit}. By assuming that there are only a small amount of salient foreground tokens and only these tokens need fine-grained computation, token compression methods can reweigh computation resources between foreground and background tokens. This can depress unnecessary computation and dramatically reduce the computational burden. However, existing token compression methods are initially designed for 2D vision tasks and conduct token compression without 3D-aware features or priors. The lack of 3D awareness leads to sub-optimal token compression when facing objects with complicated 3D motion transformations. It thus significantly hurts the performance if they are applied to multi-view 3D detectors.

To accelerate the multi-view 3D detectors with ViTs while maintaining high performance, we propose a simple yet effective method called \textbf{To}ken\textbf{C}ompression \textbf{3D} (ToC3D) in this paper, shown in Fig.~\ref{fig:intro}(a). The key insight is: the object queries from history predictions, which contain 3D motion information, can serve as the foreground prior of high quality. By leveraging these object queries, we can achieve 3D-aware token compression and foreground-oriented computing resource assignment. This insight allows us to further extend the philosophy of sparse query-based methods from the 3D decoder to the whole pipeline and achieve more efficient multi-view 3D object detection.

Specifically, ToC3D mainly consists of two designs: motion query-guided token selection strategy (MQTS) and dynamic router. MQTS takes image tokens and history object queries as inputs, models the motion information of object queries, and calculates the importance score of each image token through the attention mechanism. With the supervision of projected ground truth objects, it learns to divide image tokens into salient and redundant parts. Then, we pass them to the dynamic router for feature extraction of high efficiency, whose core is assigning more computing resources to the salient foreground proposals and removing unnecessary consumption for acceleration. After integrating these two modules with ViT, ToC3D further boosts the efficiency of sparse query-based multi-view 3D detectors and keeps their impressive performance.

We evaluate our method on the nuScenes~\cite{caesar2020nuscenes} dataset. The extensive experiments prove the effectiveness of our method, as shown in Fig.~\ref{fig:intro}(b). In detail, when compared with the StreamPETR~\cite{wang2023streampetr} baseline, our method can nearly maintain the performance with up to 30\% inference speedup, and further accelerate the baseline to the same level with other ResNet-based multi-view 3D detector~\cite{park2022solofusion} while keeping the performance superiority. The accuracy-efficiency tradeoff improvements are consistent after scaling up the ViT and input image resolution. Moreover, our method can also be applied to other baselines as well.

In summary, the main contributions of our method are two-fold: \textbf{1)} We point out that the naive usage of ViTs brings unnecessary computational burdens and strongly hinders the inference speed of sparse query-based multi-view 3D detectors. \textbf{2)} We propose a simple and efficient method called ToC3D to solve the problem, which uses history object queries with motion information to achieve 3D motion-aware token compression, and finally obtain faster ViTs.

\section{Related Work}
\subsection{Multi-view 3D Object Detection}
Multi-view 3D object detection has many advantages when deploying to the real world, given its low costs and simple sensor setups (\ie, it only needs cameras). 
Existing methods can be mainly categorized into two types: dense BEV-based paradigm~\cite{yang2023bevformerv2, li2023bevnext, li2023fastbev, li2023fbbev, zhang2023sam3d} and sparse query-based paradigm~\cite{liu2022petr, liu2023petrv2, lin2023sparse4dv2, lin2023sparse4dv3}.

For dense BEV-based paradigm, many works~\cite{huang2021bevdet, li2023bevdepth, park2022solofusion} use the explicit view transformation (\eg, LSS~\cite{philion2020lss}) to transform image features into dense BEV. BEVDet~\cite{huang2021bevdet} is the pioneer work of this paradigm. 
BEVDepth~\cite{li2023bevdepth} leverages explicit depth supervision to facilitate accurate depth estimation, and SOLOFusion~\cite{park2022solofusion} combines long-term and short-term temporal stereo for better depth estimation, both improve the performance significantly. Instead of explicit view transformation, BEVFormer~\cite{li2022bevformer} pre-defines grid-shaped BEV queries and aggregates dense BEV features through attention, which is implicit. PolarFormer~\cite{jiang2023polarformer} explores the polar coordinate system to replace the grid-shaped coordinate system. Since dense BEV-based methods need to extract dense BEV features, the computation and memory costs are relatively high. 

For the sparse query-based paradigm, DETR3D~\cite{wang2022detr3d} initializes a set of 3D queries and aggregates features by projecting 3D queries into the 2D image plane. PETR~\cite{liu2022petr} encodes the position information of 3D coordinates into image features, eliminating the need for 3D query projection. CAPE~\cite{xiong2023cape} and 3DPPE~\cite{shu20233dppe} further improve the quality of 3D position information. SparseBEV~\cite{liu2023sparsebev} introduces adaptability to the detector in both BEV and image space. For temporal 3D detection, Sparse4D~\cite{lin2022sparse4d} proposes sparse 4D sampling to aggregate features from multi-view/scale/timestamp.
StreamPETR~\cite{wang2023streampetr} introduces a memory queue to store history object queries for long-term temporal information propagation. Because these methods pass image features directly to the 3D decoder for detection, high-quality image features are beneficial. With advancing pre-trained ViTs, the sparse query-based methods~\cite{wang2023streampetr, lin2022sparse4d, liu2023sparsebev} have achieved state-of-the-art performance and nearly dominated the leaderboard. 
However, their inference speeds are mainly hindered by the backbone due to the computational burden of ViT, which motivates us to trim the ViT backbone.

\subsection{Token Compression for Vision Transformers}
Vision transformer (ViT)~\cite{dosovitskiy2020vit} has gone viral in various computer vision tasks~\cite{he2022mae,xie2021segformer} due to its strong feature extraction ability. The visualization of trained ViT shows sparse attention maps, which means the final prediction only depends on a subset of salient tokens. Based on this observation, many works~\cite{rao2021dynamicvit, bolya2022tome, xu2022evovit, kong2022spvit, long2023beyondattentivetokens} attempt to speed up ViT by removing redundant tokens, dubbed token compression. Specifically, DynamicViT~\cite{rao2021dynamicvit} introduces a lightweight prediction module to estimate the importance score of each token and then prunes redundant tokens progressively and dynamically. A-ViT~\cite{yin2022Avit} further proposes a dynamic halting mechanism.
EViT~\cite{liang2022Evit} leverages class token attention to identify the importance of tokens, then reserves the attentive image tokens and fuses the inattentive tokens. AdaViT~\cite{meng2022adavit} further prunes at attention head, and block level. Si \etal~\cite{long2023beyondattentivetokens} jointly considers the token importance and diversity. Evo-ViT~\cite{xu2022evovit} presents a self-motivated slow-fast token evolution approach, which maintains the spatial structure and information flow. All the methods are initially designed for 2D vision tasks and conduct token compression without 3D-aware priors. 

In this paper, borrowing from the methods in~\cite{rao2021dynamicvit,kong2022spvit,xu2022evovit}, we extend the token compression from the 2D domain to the 3D domain by leveraging history object queries and modeling the 3D motion information, leading to 3D motion-aware token compression tailored for 3D object detection.

\section{Method}
\subsection{Overview}
Sparse query-based methods~\cite{liu2022petr, lin2022sparse4d, wang2023streampetr} improve the efficiency of 3D detectors by mainly modeling sparse object-centric queries as the foreground proxy instead of the whole 3D scene. However, we argue that for existing sparse query-based methods, there still exists much room for efficiency improvements, as they treat foreground and background all the same in the backbone. When using ViT~\cite{li2022vitdet, fang2023eva02} to achieve extraordinary performance, the backbone becomes the bottleneck of inference speed. 

\begin{figure}[t]
\begin{center}
\includegraphics[width=0.99\linewidth]{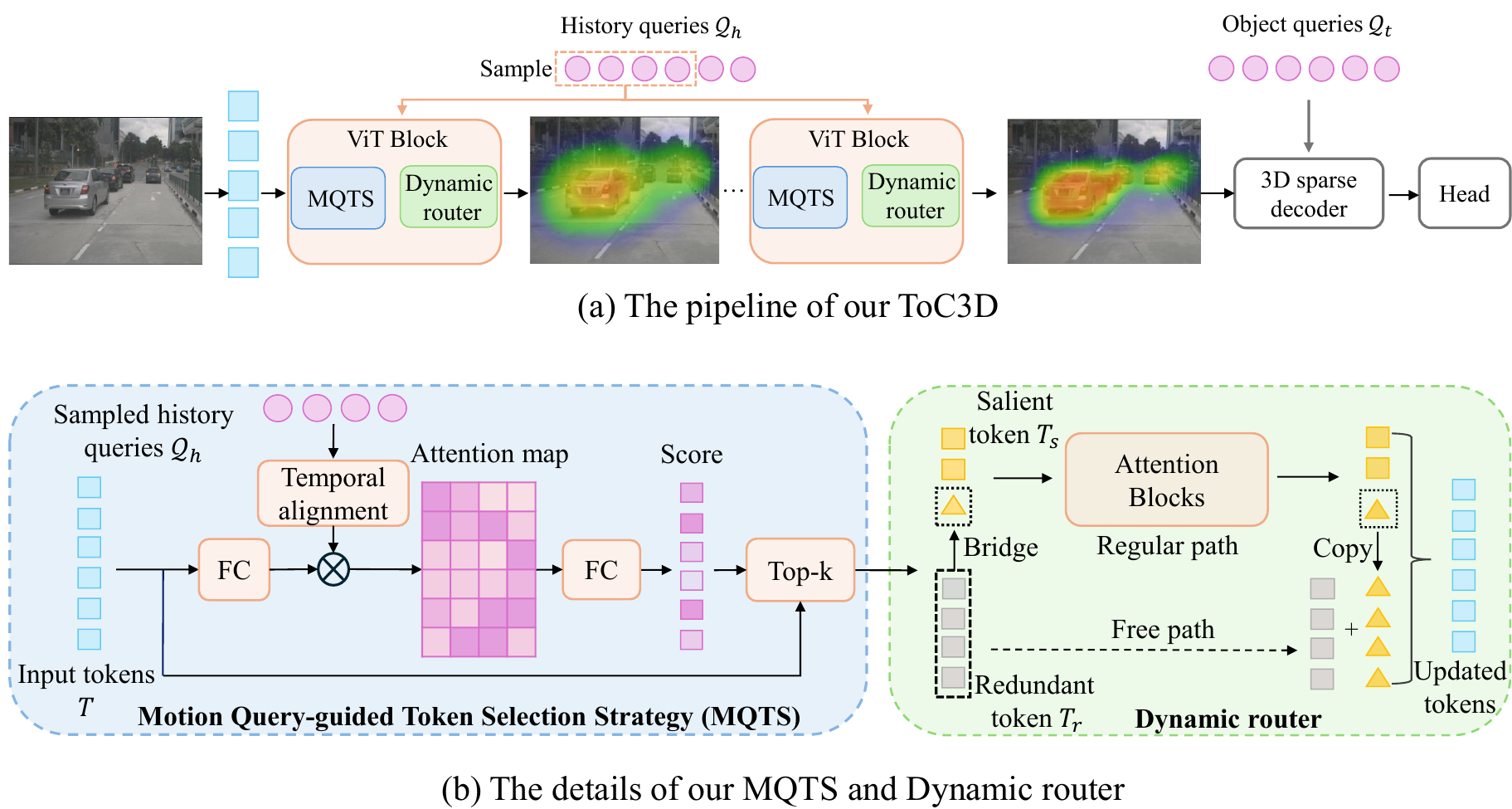}
\end{center}
\caption{
(a) The overall architecture of ToC3D, which trims each block of ViT backbone through two designs: Motion Query-guided Token Selection strategy (MQTS) and dynamic router. (b) MQTS takes motion queries from history frames as inputs, calculates the importance score, and splits image tokens into salient and redundant tokens. Dynamic router passes these tokens to different paths for efficient feature extraction.
}
\label{fig:pipeline}
\label{fig:token_select}
\end{figure}

To tackle the above problem, we propose to leverage token compression for extending the design philosophy of sparse query-based methods to the ViT backbone, named \textbf{To}ken\textbf{C}ompression\textbf{3D} (ToC3D). As Fig.~\ref{fig:pipeline}(a) shows, ToC3D mainly comprises two designs: motion query-guided token selection strategy (MQTS) and dynamic router. The token compression in each block happens as follows:\textbf{ 1)} First, MQTS takes image tokens and history object queries as inputs and calculates the importance score of each image token through the attention between image tokens and history queries, splitting image tokens into salient and redundant ones. \textbf{2)} Then, the dynamic router is used to extract features from different groups of tokens efficiently. Salient tokens are passed to the regular path, which consists of many attention blocks. The free path with the identity layer is used for redundant tokens to save computational costs. To keep the interaction between salient and redundant tokens in attention blocks, we merge redundant tokens into one bridge token and append it with salient tokens before the regular path. \textbf{3) }Finally, after obtaining features of salient and redundant tokens, we rearrange salient and redundant tokens to meet the compatibility with typical 3D object detectors.

By stacking token compression-empowered blocks, the computing resources are dynamically and more intensively assigned to the foreground proposals, which removes unnecessary consumption and accelerates the inference remarkably. Ultimately, we effectively trim the ViT backbone and develop a more efficient sparse query-based multi-view 3D detector with the 3D sparse decoder.

\subsection{Motion Query-guided Token Selection Strategy}
\label{sec: token_selection}

Motion query-guided token selection strategy (MQTS) is meant to measure the importance score of each image token and split tokens into salient/redundant tokens. Generally, salient and redundant tokens usually contain information about foreground objects and background things. Based on this, MQTS essentially segments the foreground tokens in the image, and the history object queries with 3D motion information can serve as the foreground prior of high quality, which leads to the motion query-guided token selection design shown in Fig.~\ref{fig:token_select}(b).

Specifically, MQTS first takes history query contents $Q_{h}^{c} \in \mathbb{R}^{N_q \times C_q}$, history query reference points $Q_{h}^{p} \in \mathbb{R}^{N_q \times 4}$ (in homogeneous coordinates), and image tokens of current frame $T \in \mathbb{R}^{N \times C}$ as inputs, where $N_q, N$ are the number of history queries and image tokens, $C_q, C$ are the channels of history queries and image tokens, respectively. Then, MQTS uses the following process to split tokens accurately and efficiently:

\textbf{Motion query preparation.} Since spatial transformation exists between current and history frames, we introduce the temporal alignment process to align $Q_{h}^{p}$ with the current ego-coordinate system. Following~\cite{wang2023streampetr}, we use the motion-aware layer normalization. First, we view all objects as static and align $Q_{h}^{p}$ to the current frame using the ego transformation matrix:
\begin{equation}
\label{eq: motion_1}
    \hat{Q}^{p} = E_{h} \cdot Q_{h}^{p},
\end{equation}
where $\hat{Q}^{p}$ is the aligned history query reference points and $E_{h}$ is the ego transformation matrix from history frames to current frame. Then, we model the motion of movable objects by a conditional layer normalization, with affine transformation coefficients calculated as follows:

\begin{equation}
    V_m = \mathrm{PE} \left( \left[v, \triangle t, E_{h} \right] \right), \quad
    \gamma = \mathrm{Linear}(V_m), \quad
    \beta = \mathrm{Linear}(V_m),
\end{equation}

where $V_m$ is the encoded motion vector, $\gamma, \beta$ are affine transformation coefficients used in layer normalization, $[\cdot]$ is the concatenate operator, $v$ is the velocity of queries, $\triangle t$ is the time difference between history queries and current frame. $\mathrm{PE}(\cdot)$ is the positional encoding function, and we adopt the sine-cosine encoding used in NeRF~\cite{mildenhall2021nerf}. Finally, we encode the motion information through conditional layer normalization:
\begin{equation}
\label{eq: motion_2}
    \widetilde{Q}^{p} = \gamma \cdot \mathrm{LN} \left(\mathrm{MLP} \left(\hat{Q}^{p} \right) \right) + \beta, \qquad \widetilde{Q}^{c} = \gamma \cdot \mathrm{LN} \left(Q_{h}^{c} \right) + \beta,
\end{equation}

where $\widetilde{Q}^{c} \in \mathbb{R}^{N_q \times C_q}, \widetilde{Q}^{p} \in \mathbb{R}^{N_q \times C_q}$ are temporally aligned history query contents and reference points embedding, and $\mathrm{MLP}(\cdot)$ is a multi-layer perceptron used to convert reference points into embedding. 

\textbf{Importance score calculation.}  After obtaining the temporally aligned history query, we segment salient foreground tokens. To better extract the foreground prior, we leverage the attention mechanism to calculate the importance score of each image token. We first add temporally aligned history query contents and reference points embedding to get the new query embedding and then use a linear layer to align the dimensions of image tokens and query embedding:
\begin{equation}
    \widetilde{Q} = \widetilde{Q}^{p} + \widetilde{Q}^{c}, \quad
    \widetilde{T} = \mathrm{Linear} \left( T \right),
\end{equation}
where $\widetilde{Q} \in \mathbb{R}^{N_q \times C_q}, \widetilde{T} \in \mathbb{R}^{N \times C_q}$ are dimension-aligned query embedding and image tokens respectively, $N, N_q$ are the number of image tokens and history queries, respectively. $C_q$ is the channel number of history queries. Next, we obtain the attention map through efficient matrix multiplication:
\begin{equation}
\label{eq: attention}
    A = \frac{\widetilde{T} \widetilde{Q}^\top}{\sqrt{C_q}}.
\end{equation}

Essentially, the attention map $A \in \mathbb{R}^{N \times N_q}$ models the correlations between image tokens and history queries and thus can represent the foreground information density of each token since history queries contain foreground priors. By aggregating the foreground information density via a simple linear transformation with sigmoid activation, we determine the importance score $S \in \mathbb{R}^{N \times 1}$ of each image token:
\begin{equation}
    S = \mathrm{Sigmoid} \left( \mathrm{Linear} \left( A \right) \right).
\end{equation}

Finally, we select the top-k image tokens as salient tokens according to importance score, making batch processing easier:
\begin{equation}
    T_{s} = \mathrm{Topk} \left( T, S, N_s \right),  \quad
    T_{r} = \mathrm{Topk} \left( T, -S, N_r \right),
\end{equation}
where $T_{s} \in \mathbb{R}^{N_s \times C}$ is the salient tokens and $T_{r} \in \mathbb{R}^{N_r \times C}$ is the redundant tokens. $N_s = \rho N, N_r = \left(1-\rho \right) N$ are the number of salient and redundant tokens respectively. $\rho$ is the pre-set constant keeping ratio of each block. Increasing $\rho$ will scale up the number of salient tokens; otherwise, scale up the number of redundant tokens, and thus, we can control the inference speed by adjusting $\rho$.

Although the importance score calculation is light-weight, it still brings some overhead. We found that updating the importance score of every layer does not bring noticeable improvements, so we only update the importance score before specific transformer layers for efficiency, and the layers in between will reuse the newest importance score.

\textbf{History query sampling.} Although history object queries carry high-quality foreground priors, we empirically found that not all history object queries are valuable. This is because the number of object queries is larger than that of objects of interest in a typical DETR-style detector, and many object queries do not correspond to foreground objects but background things. If we directly use all these history queries in MQTS, the importance score calculation will be biased by background information. Thanks to the confidence measurement of object queries, we can simply solve this problem by sampling history queries according to their confidence scores, which the decoder has already predicted in the history frames. In detail, we sort the history queries according to their confidence scores and select the top-$N_q$ queries as inputs of MQTS.

\subsection{Dynamic Router}
After splitting tokens into salient and redundant ones, we introduce a dynamic router to accelerate the inference speed while keeping information losses as low as possible, shown in Fig.~\ref{fig:pipeline}(b). Considering that not all redundant tokens correspond to the background, these tokens may contain some potential information for detection, and the information can be passed to foreground tokens via attention interactions. We merge redundant tokens into a single bridge token and append it with salient tokens, allowing interactions between salient and redundant tokens through the bridge token. Then, we use more neural layers (\ie, regular path) to extract rich semantic and geometric information in salient tokens, while using shallow layers (\ie, free path, we use the identity layer in this paper) to keep the information in redundant tokens. 

Formally, we use the importance scores to conduct weighted sum with redundant tokens for getting bridge token:
\begin{equation}
\label{eq: repre_begin}
    T_b = \frac{\sum_{i=1}^{N_r} S_{r:i} T_{r:i}}{\sum_{i=1}^{N_r} S_{r:i}}
\end{equation}
where $T_{b} \in \mathbb{R}^{1 \times C}$ is the bridge token, $S_{r:i}$ is the importance score of the $i$-th redundant token, and $T_{r:i}$ is the $i$-th redundant token.

Afterwards, we append the bridge token after salient tokens and pass them to the regular path for extracting rich semantic and geometric features:
\begin{equation}
    \left[T_s^{\prime}, T_b ^{\prime}\right] = \mathrm{Blocks} \left( \left[ T_{s}, T_{b}\right] \right),
\end{equation}
where $T_s^{\prime}, T_b ^{\prime}$ are updated salient and bridge tokens, $\mathrm{Blocks}$ are transformer encoder blocks, typically consisting of several window attention layers and a global attention layer for multi-view 3D object detection.

For redundant tokens, we pass them to the fast path (\ie, identity layer) and add them with the updated bridge token:
\begin{equation}
\label{eq: repre_end}
    T_r^{\prime} = T_{r} + \mathrm{Repeat}(T_b^{\prime}, N_r),
\end{equation}
where $T_r^{\prime}$ is the updated redundant tokens and $\mathrm{Repeat}(x, y)$ repeats $x$ by $y$ times.

Finally, we combine the updated salient and redundant tokens to obtain the updated image tokens. Thanks to this simple yet effective dynamic router, we refine image tokens more efficiently and meet the compatibility with typical multi-view 3D detectors.

\section{Experiments}
\subsection{Dataset and Metrics}
We evaluate our method on the large-scale nuScenes~\cite{caesar2020nuscenes} dataset, consisting of 700 scenes for training, 150 for validation, and 150 for testing. The data of each scene is captured by six cameras at 10Hz, with full 360$^\circ$ field of view (FOV). We use annotations of 10 classes: car, truck, construction vehicle, bus, trailer, barrier, motorcycle, bicycle, pedestrian, and traffic cone.
We use the official nuScenes metrics for comparison: the nuScenes detection score (NDS), the mean average precision (mAP), the average translation error (ATE), the average scale error (ASE), the average orientation error (AOE), the average velocity error (AVE), average attribute error (AAE). 

\subsection{Implementation Details}
We select recently representative StreamPETR~\cite{wang2023streampetr} as our basic pipeline, considering its high performance. 
For the backbone, we adopt ViT-B, ViT-L~\cite{dosovitskiy2020vit} and conduct token compression on them. We use the Gaussian Focal Loss~\cite{law2018cornernet} to supervise the MQTS, with the ground truth coming from projected bounding boxes. The model is trained on 8 NVIDIA V100 with a total batch size of 16 for 24 epochs. The inference speed is tested on a single RTX3090. AdamW~\cite{loshchilov2018adamw} is used as the optimizer. The augmentation follow the StreamPETR~\cite{wang2023streampetr}, and without CBGS~\cite{zhu2019cbgs}. The detailed configurations can be viewed in Tab.~\ref{tab:implementation}.

\begin{table}[t]
\setlength{\tabcolsep}{1.1mm}
\centering
\scriptsize
\caption{Details of settings. }
\label{tab:implementation}

\begin{tabular}{ lcccc }
\toprule
Configurations & \multicolumn{2}{c}{ToC3D-Fast} & \multicolumn{2}{c}{ToC3D-Faster}\\
\midrule
Backbone & ViT-B & ViT-L & ViT-B & ViT-L \\
Dim. of image token $C$ & 768 & 1024 & 768 & 1024\\
Num. of layers & 12 &  24  & 12 & 24 \\
Num. of object query $N_q$ & 64 & 64 & 64 & 64\\
Dim. of object query $C_q$ & 256 & 256 & 256 & 256\\
Keeping ratios $\rho$ & 0.7, 0.5, 0.5 & 0.7, 0.5, 0.5 & 0.5, 0.4, 0.3 & 0.5, 0.4, 0.3 \\
Loc. of importance score $S$ updating & 3, 6, 9 & 6, 12, 18 & 3, 6, 9 & 6, 12, 18 \\
Token compression loss weight & 5.0 & 5.0 & 5.0 & 5.0 \\
Pretrained weight & SAM~\cite{kirillov2023sam} & EVA-02~\cite{fang2023eva02} & SAM~\cite{kirillov2023sam} & EVA-02~\cite{fang2023eva02} \\
\bottomrule

\end{tabular}
\end{table}

\begin{table}[t]
\setlength{\tabcolsep}{0.7mm}
\centering
\scriptsize
\caption{The main results on the nuScenes val set. We report the backbone inference time (before the slash) and the whole pipeline inference time (after the slash) to illustrate the impact of efficient backbone better. $^\dag$means using larger image resolution.}
\label{tab:main_result_val}

\scalebox{0.765}{
\begin{tabular}{ lccccccccc }
\toprule
Method & Backbone & NDS(\%)$\uparrow$ & mAP(\%)$\uparrow$ & mATE$\downarrow$ & mASE$\downarrow$ & mAOE$\downarrow$ & mAVE$\downarrow$ & mAAE$\downarrow$ & Infer. Time (ms)$\downarrow$\\
\midrule
BEVDet~\cite{huang2021bevdet} & R50 & 37.9 & 29.8 & 0.725 & 0.279 & 0.589 & 0.860 & 0.245 & - / 59.9 \\
BEVDepth~\cite{li2023bevdepth} & R50 & 47.5 & 35.1 & 0.639 & 0.267 & 0.479 & 0.428 & 0.198 & - / 63.7 \\
SOLOFusion~\cite{park2022solofusion} & R50 & 53.4 & 42.7 & 0.567 & 0.274 & 0.511 & 0.252 & 0.181 & - / 87.7 \\
StreamPETR~\cite{wang2023streampetr} & ViT-B & 55.4 & 45.8 & 0.608 & 0.272 & 0.415 & 0.261 & 0.191 & 85.2 / 115.0\\
\rowcolor{gray!20} Ours-Fast & ViT-B & 55.2 & 46.0 & 0.604 & 0.270 & 0.449 & 0.261 & 0.196 & 70.3{\ddplus{-17.5\%}} / 94.0{\ddplus{-18.3\%}}\\
\rowcolor{gray!20} Ours-Faster & ViT-B & 54.9 & 45.3 & 0.594 & 0.271 & 0.443 & 0.258 & 0.207 & 59.2{\ddplus{-30.5\%}} / 85.0{\ddplus{-26.1\%}}\\
\midrule
DETR3D$^\dag$~\cite{wang2022detr3d} & R101 & 43.4 & 34.9 & 0.716 & 0.268 & 0.379 & 0.842 & 0.200 & - / 270.3 \\
BEVFormer$^\dag$~\cite{li2022bevformer} & R101 & 51.7 & 41.6 & 0.673 & 0.274 & 0.372 & 0.394 & 0.198 & - / 333.3 \\
Sparse4D$^\dag$~\cite{lin2022sparse4d} & R101 & 54.1 & 43.6 & 0.633 & 0.279 & 0.363 & 0.317 & 0.177 & - / 232.6\\
StreamPETR~\cite{wang2023streampetr} & ViT-L & 61.2 & 52.1 & 0.552 & 0.251 & 0.249 & 0.237 & 0.196 & 290.0 / 317.0 \\
\rowcolor{gray!20} Ours-Fast & ViT-L & 60.9 & 51.7 & 0.552 & 0.250 & 0.268 & 0.229 & 0.195 & 253.0{\ddplus{-12.8\%}} / 281.0{\ddplus{-11.4\%}}\\
\rowcolor{gray!20} Ours-Faster & ViT-L & 60.5 & 51.3 & 0.562 & 0.250 & 0.265 & 0.230 & 0.203 & 209.0{\ddplus{-28.0\%}} / 237.2{\ddplus{-25.2\%}}\\
\midrule
StreamPETR$^\dag$~\cite{wang2023streampetr} & ViT-L & 62.7 & 55.8 & 0.552 & 0.256 & 0.287 & 0.225 & 0.201 & 1222.4 / 1309.9\\
\rowcolor{gray!20} Ours-Fast$^\dag$ & ViT-L & 62.6 & 54.9 & 0.536 & 0.254 & 0.259 & 0.230 & 0.206 & 964.8{\ddplus{-21.1\%}} / 1051.9{\ddplus{-19.7\%}}\\
\rowcolor{gray!20} Ours-Faster$^\dag$ & ViT-L & 61.9 & 54.3 & 0.560 & 0.257 & 0.230 & 0.234 & 0.201 & 791.0{\ddplus{-35.3\%}} / 878.5{\ddplus{-33.0\%}}\\
\bottomrule

\end{tabular}
}

\end{table}

\subsection{Main Results}
We compare our method with the basic pipeline StreamPETR~\cite{wang2023streampetr} and other popular multi-view 3D detectors on nuScenes~\cite{caesar2020nuscenes} val set.

The main results are illustrated in Tab.~\ref{tab:main_result_val}. When using ViT-B as the backbone, our method (ToC3D-Fast) can perfectly maintain NDS and mAP compared with the StreamPETR method, with nearly 20\% speedup. If marginal 0.5\% NDS and mAP drop are allowed, our method (ToC3D-Faster) can further accelerate the backbone by 30\% and the whole pipeline by 26\%. Notably, ToC3D-Faster performs at the same level as StreamPETR while only costs like SOLOFusion with R50 backbone, indicating the effectiveness of our method. 

When using ViT-L as the backbone, our method (ToC3D-Fast) achieves nearly lossless performance compared to the basic StreamPETR while accelerating the whole pipeline by 36ms. Furthermore, with a performance loss of no more than 0.9\%, our method (ToC3D-Faster) brings 25\% inference speed gains and runs at the same speed with Sparse4D~\cite{lin2022sparse4d} while keeping the vast performance superiority (\ie, over 6.4\% NDS and 7.7\% mAP). 

Furthermore, scaling up the input image resolution to 800$\times$1600, our method can tremendously reduce the inference time by 258ms and 431ms with ToC3D-Fast and ToC3D-Faster settings, saving a considerable amount of computing resources for detection deployed on the cloud. The results prove that our method can achieve better trade-offs and greatly improve the efficiency of 3D detectors.

\begin{table}[t]
\setlength{\tabcolsep}{1.5mm}
\centering
\scriptsize
\caption{Results of comparison between our method with different 2D token compression methods on the nuScenes val set. We apply these methods to the StreamPETR baseline and have carefully tuned their hyper-parameters to achieve their best results.}
\label{tab:ablation_score_type}
\scalebox{.88}{
\begin{tabular}{ lccccc }
\toprule
Method & Type & Keeping Ratio $\rho$ &NDS(\%)$\uparrow$ & mAP(\%)$\uparrow$ & Infer. Time (ms)$\downarrow$\\
\midrule
StreamPETR~\cite{wang2023streampetr} & - & - & 61.2 & 52.1 & 290.0 / 317.0\\
\midrule
+ Random & Random & 0.7, 0.5, 0.5 & 56.7 & 46.5 & 250.1 / 277.9\\
+ Random & Random & 0.5, 0.4, 0.3 & 48.5 & 36.0 & 207.3 / 235.1\\
\midrule
+ DynamicViT~\cite{rao2021dynamicvit} & Score-based & 0.7, 0.5, 0.5 & 59.7 & 50.5 & 249.8 / 277.4\\
+ DynamicViT~\cite{rao2021dynamicvit} & Score-based & 0.5, 0.4, 0.3 & 59.3 & 49.3 & 208.0 / 233.4\\
\midrule
+ SparseDETR~\cite{roh2021sparsedetr} & Score-based & 0.7, 0.5, 0.5 & 59.3 & 49.7 & 249.0 / 280.5\\
+ SparseDETR~\cite{roh2021sparsedetr} & Score-based & 0.5, 0.4, 0.3 & 59.1 & 49.2 & 208.5 / 236.5\\
\midrule
\rowcolor{gray!20}Ours-Fast & Motion Query-guided & 0.7, 0.5, 0.5 & 61.0  & 52.3& 253.0 / 281.0 \\
\rowcolor{gray!20}Ours-Faster & Motion Query-guided & 0.5, 0.4, 0.3 & 60.3 & 51.2 & 209.0 / 237.2 \\
\bottomrule

\end{tabular}
}

\end{table}

\subsection{Analysis}
We conduct experiments for analysis of our method using ViT-L as the backbone. All models are trained for only 12 epochs and evaluated on the val set.

\textbf{Compared to 2D token compression methods.} 
To prove the effectiveness of our motion query-guided token selection strategy (MQTS), we compare our method with typical 2D token compression methods DynamicViT~\cite{rao2021dynamicvit} and SparseDETR~\cite{roh2021sparsedetr}. For a fair comparison, we replace the MQTS with these two methods, keep the dynamic route unchanged, and tune these methods to their best performance. We also compare with the Random token compression.

As listed in Tab.~\ref{tab:ablation_score_type}, it clearly shows that the random token compression brings a significant performance drop, especially when keeping ratios are low. This is because the random compression cannot capture the importance of image tokens and thus drops much helpful information. When using the score-based 2D token compression methods, the performance drop is much smaller than random compression since they are better aware of important foreground tokens and thus suffer less information loss. However, because these 2D methods only take image tokens as input, they conduct token compression without any 3D-aware features or priors. The lack of 3D awareness leads to sub-optimal token compression and thus hurts the performance severely (\ie, about 2\% mAP and 2\% NDS). 

When it comes to our method, because MQTS has history object queries as inputs, it can model the 3D motion information of objects and aggregate the rich 3D foreground priors of high quality, leading to remarkably better results than 2D competitors (more than 2\% mAP and 1.2\% NDS improvement). Notably, With efficient MQTS, our method is able to almost maintain the performance of the basic pipeline at the same speed level as 2D token compression methods, indicating the superiority of MQTS.

\textbf{Effectiveness of components.} After proving our key insight that the object queries from history predictions can serve as the foreground prior of high quality, we now study what makes this insight work. We take our method with the Faster setting as the baseline of this experiment, and we remove one component each time to measure its effectiveness, shown in Tab.~\ref{tab:ablation_motion}. It is worth noting that removing any components only slightly reduces inference time, showing the high efficiency of each component.

For setting (a), we replace the attention in Eq.~\ref{eq: attention} with a lightweight module, which brings 0.8\% mAP and 0.4\% NDS drop. This is because the attention map naturally models the correlations between image tokens and history queries and thus more explicitly represents the foreground information density of each token, leading to better importance measurement. 

For setting (b), we discard processes from Eq.~\ref{eq: motion_1}$\sim$\ref{eq: motion_2}. The degraded performance (1.1\% mAP and 0.6\% NDS) shows the importance of motion information, which adaptively handles movable objects and suppresses the noise brought by misalignment between history objects and the current coordinate system. 

For setting (c), we remove the history query sampling and use all history queries instead. This choice reduces mAP by 0.8\% and NDS by 0.4\%, showing the necessity of history query sampling, as it removes object queries corresponding to background things and prevents importance score calculation from being biased. 

For setting (d), we do not use the bridge token in the dynamic router. Because of the absence of interaction between salient and redundant tokens, potential information contained in redundant tokens can not be passed to foreground tokens, leading to information loss and is ultimately reflected in mAP and NDS.

\begin{table}[t]
\setlength{\tabcolsep}{1.5mm}
\centering
\scriptsize
\caption{Effectiveness of different components on the nuScenes val set. \textbf{Attn.} means calculating importance score through the attention mechanism. \textbf{Motion} means using motion vector encoding. \textbf{Samp. Q.} means using the sampled history queries as the inputs of MQTS. \textbf{Bri. T.} means using the bridge token in the dynamic router.}

\label{tab:ablation_motion}

\begin{tabular}{ cccccccc }
\toprule

Setting & Attn. & Motion & Samp. Q. & Bri. T. & NDS(\%)$\uparrow$ & mAP(\%)$\uparrow$ & Infer. Time (ms)$\downarrow$\\
\midrule
\rowcolor{gray!20}Ours & $\checkmark$ & $\checkmark$ & $\checkmark$ & $\checkmark$ & 60.3& 51.2  & 209.0 / 237.2 \\
\midrule
(a) & & $\checkmark$ & $\checkmark$ & $\checkmark$ &59.9  &50.4  & 206.9 / 235.2\\
(b) & $\checkmark$ &  & $\checkmark$ & $\checkmark$ & 59.7 & 50.1 & 206.4 / 234.5\\
(c) & $\checkmark$ & $\checkmark$ & & $\checkmark$ & 59.9 & 50.4 & 209.3 / 237.3\\
(d)  & $\checkmark$ & $\checkmark$ & $\checkmark$ & & 60.2 & 50.5 & 203.1 / 232.7\\
\bottomrule

\end{tabular}
\end{table}

\hspace{-28pt}
\begin{figure}[t]
\begin{minipage}[t]{0.3\textwidth}
\makeatletter\def\@captype{table}
\centering
\caption{Effect of $N_q$ on the nuScenes val set. }
\renewcommand{\arraystretch}{1.13}
\setlength{\tabcolsep}{2.mm}
\tiny
\begin{tabular}{ ccc }
\toprule
$N_q$ & NDS(\%) & mAP(\%) \\
\midrule
16 & 60.1 & 50.8 \\
32 & 60.3 & 51.0 \\
\textbf{64} & \textbf{60.3} & \textbf{51.2} \\
128 & 60.1 & 50.6 \\
256 & 59.9 & 50.4 \\
\bottomrule
\end{tabular}
\label{tab:ablation_Nq}
\end{minipage}
\setlength{\tabcolsep}{1.5mm}
\begin{minipage}[t]{0.7\textwidth}
\makeatletter\def\@captype{table}
\caption{Effect of keeping ratios on the nuScenes val set. }
\tiny
\scalebox{1.}{
\begin{tabular}{ cccc }
\toprule
$\rho$ & NDS(\%)&mAP(\%) & Infer. Time (ms)\\
\midrule
 StreamPETR~\cite{wang2023streampetr} & 61.2 & 52.1 & 290.0 / 317.0\\
\midrule
 0.7, 0.5, 0.5 & 61.0 & 52.3 & 253.0{\ddplus{-12.8\%}} / 281.0{\ddplus{-11.4\%}}\\
 0.7, 0.5, 0.3 & 60.7 & 51.6 & 235.9{\ddplus{-18.7\%}} / 264.6{\ddplus{-16.5\%}}\\
 0.5, 0.4, 0.3 & 60.3 & 51.2 & 209.0{\ddplus{-28.0\%}} / 237.2{\ddplus{-25.2\%}}\\
 0.4, 0.3, 0.2 & 59.9 & 50.5 & 185.8{\ddplus{-36.0\%}} / 217.0{\ddplus{-31.5\%}} \\
 0.4, 0.3, 0.1 & 59.8 & 50.4 & 172.1{\ddplus{-40.7\%}} / 199.0{\ddplus{-37.2\%}} \\
 0.3, 0.2, 0.1 & 59.0 & 49.1 & 155.8{\ddplus{-46.3\%}} / 183.3{\ddplus{-42.2\%}}\\
\bottomrule
\end{tabular}}
\label{tab:ablation_ratio}
\end{minipage}
\end{figure}

\textbf{Impact of history query num $N_q$.} Since we sample $N_q$ history queries as inputs of MQTS, we study the impact of different $N_q$ in this section, shown in Tab.~\ref{tab:ablation_Nq}. When $N_q=256$, we use all history queries. Otherwise, we sample top-$N_q$ history queries according to their confidence scores from history predictions. The results show that sampling history queries is always beneficial, as many history queries correspond to background things. History query sampling helps prevent the importance score from being biased by background and thus improves the performance. However, if $N_q$ is too small, it will drop many foreground queries, suffer information loss, and hurt the performance. We empirically find that $N_q = 64$ can achieve better performance.

\begin{figure}[t]
\begin{center}
\includegraphics[width=0.96\linewidth]{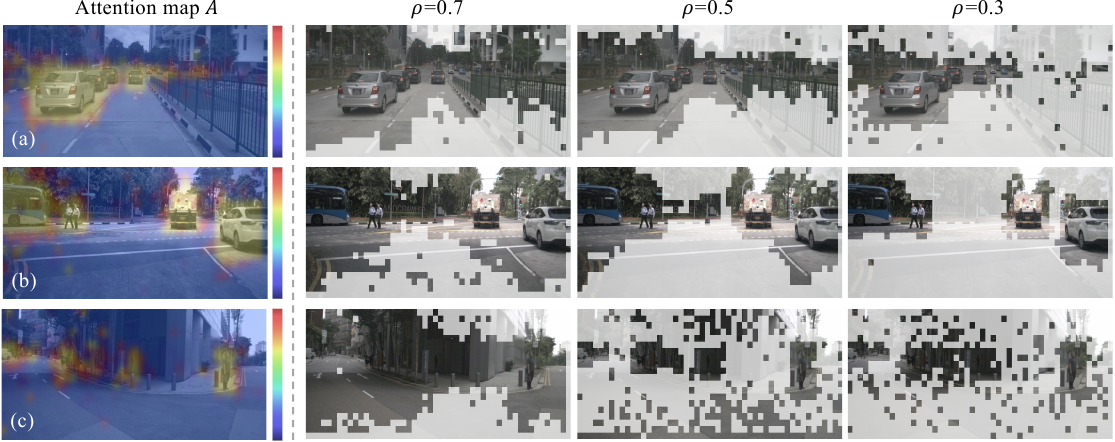}
\end{center}

\caption{The visualization of our method (better viewed in color). We visualize the attention map in importance score calculation on the left and the salient/redundant tokens after the top-k selection on the right. Redundant tokens are illustrated as translucent.}
\label{fig:vis}

\end{figure}

\begin{table}[h]
\setlength{\tabcolsep}{0.8mm}
\centering
\scriptsize

\caption{Analysis of generalization on the nuScenes val set. We report the backbone inference time before the slash and the whole pipeline inference time after the slash.}

\label{tab:generalization}
\begin{tabular}{ lccccc }
\toprule
Method & Backbone & Compression & NDS(\%) &mAP(\%) & Infer. Time (ms)\\
\midrule
StreamPETR~\cite{wang2023streampetr} & ViT-L & - & 61.2 & 52.1 & 290.0 / 317.0\\
Ours-Faster & ViT-L & $\checkmark$ & 60.3 & 51.2 & 209.0{\ddplus{-28.0\%}} / 237.2{\ddplus{-25.2\%}}\\
\midrule
Sparse4Dv2~\cite{lin2023sparse4dv2} & ViT-L & - & 58.8 & 50.9 & 278.8 / 322.0\\
Ours-Faster & ViT-L & $\checkmark$ & 58.8 & 50.1 & 206.6{\ddplus{-25.9\%}} / 244.8{\ddplus{-24.0\%}}\\
\bottomrule
\end{tabular}

\end{table}

\textbf{Impact of keeping ratios $\rho$.} Keeping ratios $\rho$ controls the number of salient tokens and decides the accuracy-speed trade-off. In this section, we conduct experiments to figure out how keeping ratios affects the efficiency of our method. From Tab.~\ref{tab:ablation_ratio}, we can get the following phenomenon: (1) We can speed up the pipeline by nearly 15\% without noticeable performance loss, and 25\% with marginal NDS loss (0.9\%). (2) In a certain range, the performance drop is nearly linear with the inference time drop (about 0.2\% NDS drop for 20ms inference time). (3) Too small keeping ratios (\ie, 0.3, 0.2, 0.1) will bring a significant performance drop. This is because foreground tokens usually account for more than 10\% image tokens, and too small keeping ratios inevitably discard foreground tokens, bringing information loss. Considering the requirements of real applications, we set the model with keeping ratios of 0.7, 0.5, 0.5 as ToC3D-Fast, and 0.5, 0.4, 0.3 as ToC3D-Faster versions, respectively, as these models have relatively better trade-off.

\textbf{Generalization.} We select StreamPETR~\cite{wang2023streampetr} as our basic pipeline, 
but this does NOT mean that the application of our method is limited. In fact, ToC3D can serve as a plug-and-play method, and we show the generalization ability of our method by applying it to another strong pipeline, Sparse4Dv2~\cite{lin2023sparse4dv2}.
The results are shown in Tab.~\ref{tab:generalization}. It indicates that the behavior of our method is consistent across different baseline methods. With the same speed-up ratio as on StreamPETR, our approach keeps the performance loss within 0.8\% and surprisingly maintains the exact NDS compared to the strong Sparse4Dv2~\cite{lin2023sparse4dv2}, proving the feasibility and effectiveness of our method on other pipelines.

\subsection{Qualitative Results}
To better study the behavior of MQTS, we visualize the attention map and salient tokens in Fig.~\ref{fig:vis}. The attention maps clearly show that our method focuses on foreground objects of interest precisely, no matter the large objects (\eg, cars and trucks in sample (a) or the small objects (\eg, pedestrians in sample (b) and (c)). This is an intuitive proof of our claim in Sec.~\ref{sec: token_selection}, \ie, the attention map models the correlations between image tokens and history queries, and thus can represent the foreground information density of each token since history queries contain foreground priors. With the 3D foreground object-aware attention map, the whole model can be more concentrated on foreground tokens when keeping ratios $\rho$ getting lower, improving the efficiency.

\subsection{Limitations}

Since our method leverages history object queries as high-quality foreground priors, we assume the inputs are temporal image sequences with contiguous information. Although this assumption narrows the application of our method, we argue that this assumption is not strong as the perception system runs at typical frequencies and perceives environments contiguously for real-world autonomous driving. The second limitation of our method is that we need to re-train the token compression model if keeping ratios are changed. Using dynamic keeping ratios with some technical tricks for stability when training would help to partially solve this limitation, which is left for our future work.

\section{Conclusion}

In this paper, we claim that the naive usage of ViTs brings unnecessary computational burden and strongly hinders the speed of existing sparse query-based multi-view 3D detectors. 
To obtain a more efficient sparse multi-view 3D detector, we propose a simple yet effective method called ToC3D. Equipped with MQTS and dynamic router, ToC3D leverages history object queries as foreground priors of high quality, models 3D motion information in them, and weighs more computing resources to important foreground tokens while compressing the information loss.
By doing so, we extend the design philosophy of sparse query-based methods from the 3D decoder to the whole pipeline. The experiments on the large-scale nuScenes dataset show that our method can boost the inference speed with marginal performance loss, and using history object queries brings better results. We hope this paper can inspire the research of efficient multi-view 3D detectors and serve as a strong baseline.

\newpage

\bibliographystyle{splncs04}
\bibliography{main}

\newpage
\appendix
\section*{Supplementary Material of ToC3D}

\section{Overview}
The supplementary material is organized as follows:
\begin{itemize}
    \item In Sec.~\ref{sec: supp_implement}, we introduce more of our method's implementation details, including the implementation details of Sparse4Dv2~\cite{lin2023sparse4dv2} version.

    \item In Sec.~\ref{sec: supp_exp}, we report more experiment results, including the ablation study of the location of importance score updating, and the performance on the nuScenes test set. This section shows more about the properties of our model.

    \item In Sec.~\ref{sec: supp_vis}, we illustrate more visualizations of our method for an intuitive understanding, which shows the effectiveness of our method in various scenes.
\end{itemize}

\section{More Implementation Details}
\label{sec: supp_implement}

\begin{table}[h]
\setlength{\tabcolsep}{4mm}
\centering
\scriptsize
\caption{Details of settings using StreamPETR~\cite{wang2023streampetr} as the basic pipeline. }

\label{tab: supp_implementation}

\begin{tabular}{ lcccc }
\toprule
Configurations & \multicolumn{2}{c}{ToC3D-Fast} & \multicolumn{2}{c}{ToC3D-Faster}\\
\midrule
Backbone & ViT-B & ViT-L & ViT-B & ViT-L \\
Num. of attention heads & 12 & 16 & 12 & 16 \\
Patch size & 16 & 16 & 16 & 16 \\
Window size of window attention layer & 14 & 16 & 14 & 16 \\
Window size of global attention layer & - & 20 & - & 20 \\
Drop path rate & 0.0 & 0.3 & 0.0 & 0.3\\
Weight decay & 0.01 & 0.01 & 0.01 & 0.01\\
Grad clip & 35 & 35 & 35 & 35 \\
Num. of warmup iterations & 500 & 500 & 500 & 500\\
\bottomrule

\end{tabular}

\end{table}

\begin{table}[h]
\setlength{\tabcolsep}{4.0mm}
\centering
\scriptsize
\caption{Details of settings using Sparse4Dv2~\cite{lin2023sparse4dv2} as basic pipeline. }

\label{tab: supp_implementation_sparse4d}

\begin{tabular}{ lcc }
\toprule
Configurations & ToC3D-Fast & ToC3D-Faster\\
\midrule
Backbone &  ViT-L  & ViT-L \\
Keeping ratios $\rho$ & 0.7, 0.5, 0.5 & 0.5, 0.4, 0.3 \\
Token compression loss weight & 1.0 & 1.0 \\
Learning rate of backbone & 2.5e-5 & 2.5e-5 \\
Weight decay & 0.001 & 0.001\\
Grad clip & 25 & 25\\
\bottomrule

\end{tabular}

\end{table}

We report more detailed implementation settings in Tab.~\ref{tab: supp_implementation}, including the detailed architecture of ViTs and more training hyper-parameters.

To prove the generalization of our method, we evaluate our method on the Sparse4Dv2~\cite{lin2023sparse4dv2}. The detailed settings are listed in Tab.~\ref{tab: supp_implementation_sparse4d}, and we only report settings different from that of the StreamPETR version for simplicity. For data augmentation, we follow the official settings.

\section{Experiment Results}
\label{sec: supp_exp}

\begin{table}[t]

\centering
\scriptsize
\caption{Effect of location of importance score updating on the nuScenes val set.  We report the backbone inference time before the slash and the whole pipeline inference time after the slash.}

\label{tab: supp_ablation_loc}
\scalebox{1.}{
\begin{tabular}{ cccc }
\toprule
Location & NDS(\%)&mAP(\%) & Infer. Time (ms)\\
\midrule
 StreamPETR~\cite{wang2023streampetr} & 61.2 & 52.1 & 290.0 / 317.0\\
\midrule
 4, 10, 16 & 60.2 & 50.9 & 205.0 {\ddplus{-29.3\%}}/ 233.3{\ddplus{-26.4\%}}\\
 6, 12, 18 & 60.3 & 51.2 & 209.0{\ddplus{-28.0\%}} / 237.2{\ddplus{-25.2\%}}\\
 7, 13, 19 & 60.3 & 51.1 & 217.7{\ddplus{-24.9\%}} / 248.7{\ddplus{-21.5\%}}\\
 9, 15, 21 & 60.5 & 51.4 & 228.6{\ddplus{-21.2\%}} / 256.2{\ddplus{-19.2\%}} \\
 10, 16, 22 & 61.0 & 52.2 & 235.8{\ddplus{-18.7\%}} / 264.0{\ddplus{-16.7\%}} \\
\bottomrule
\end{tabular}}
\vspace{10pt}

\setlength{\tabcolsep}{0.2mm}
\centering
\scriptsize
\caption{We use StreamPETR~\cite{wang2023streampetr} as our baseline and list the performance on the nuScenes test set.  We report the backbone inference time before the slash and the whole pipeline inference time after the slash.}

\label{tab: supp_main_result_test}
\scalebox{0.765}{
\begin{tabular}{ lcccccccccc }
\toprule
Method & Backbone & Resolution & NDS(\%)$\uparrow$ & mAP(\%)$\uparrow$ & mATE$\downarrow$ & mASE$\downarrow$ & mAOE$\downarrow$ & mAVE$\downarrow$ & mAAE$\downarrow$ & Infer. Time (ms)$\downarrow$\\
\midrule
StreamPETR~\cite{wang2023streampetr} & ViT-L & 320 $\times$ 800 & 62.9 & 55.2 & 0.504 & 0.246 & 0.333 & 0.261 & 0.125 & 290.0 / 317.0 \\

\rowcolor{gray!20} Ours-Faster & ViT-L & 320 $\times$ 800 & 62.6 & 54.2 & 0.489 & 0.246 & 0.330 & 0.269 & 0.117 & 209.0{\ddplus{-28.0\%}} / 237.2{\ddplus{-25.2\%}}\\
\bottomrule

\end{tabular}
}
\end{table}

\subsection{Impact of Loc. of Importance Score $S$ Updating}
Since we only update the importance score $S$ before specific transformer layers, we study the impact of these locations in this section. We conduct experiments using ToC3D-Faster with ViT-L backbone, and all models are trained for 12 epochs. The results are listed in Tab.~\ref{tab: supp_ablation_loc}.

We can see that the location of importance score updating affects the accuracy-speed tradeoff. When we conduct token compression in the deeper layers, the NDS and mAP are higher with the sacrifice of inference speed. It is worth noting that when updating the importance score at the 10th, 16th, and 22nd layers, the performance is competitive with the StreamPETR~\cite{wang2023streampetr} baseline with about 17\% acceleration. We empirically find that updating the importance score at the 6th, 12th, and 18th layers can achieve a better tradeoff, which is set by default.

\subsection{Performance on Test Set}
We report the performance of ToC3D on the nuScenes~\cite{caesar2020nuscenes} test set in this section. We train all methods on the train and val set for 24 epochs, and then send the inference results of the test set to the official server for evaluation. We test with input resolution as 320 $\times$ 800. 

Tab.~\ref{tab: supp_main_result_test} shows that the performance is consistent with that on nuScenes val set, \ie, with a performance loss of no more than 0.3\% NDS and 1.0\% mAP, our method (ToC3D-Faster) brings 25\% inference speed gains. Interestingly, our method can achieve comparable or even better performance when it comes to the detailed metrics (\ie, mATE, mASE, mAOE, and mAAE). This phenomenon is aligned with our foreground-centric design, as our method weighs more computation resources to foreground tokens, and the model is more object-aware.

\section{Profiling Analysis}

\begin{table}[t] 
\setlength{\tabcolsep}{1.0mm}
\centering
\scriptsize
\caption{The profiling analysis.}
\label{tab:profile}
\begin{tabular}{ cccccc}
\toprule
Method & Module & MACs (G) & FLOPs (G) & Memory (MB) & Time (ms) \\
\midrule

\multirow{4}{*}{\tabincell{c}{StreamPETR\\(Baseline)}} & Backbone & 2280.0 & 4560.0 & 4972.5 & 290.0 \\
 & Decoder  & 13.3 & 26.6 & 405.0 & 22.6 \\
 & Head & 7.2 & 14.4 & 94.4 & 4.4 \\
\rowcolor{gray!20} \cellcolor{white}  & \textbf{Total}  & 2300.5 & 4601.0 & 5471.9 & 317.0\\

\cmidrule{2-6}
\multirow{4}{*}{Ours-Faster} & Backbone  & 1545.0 & 3090.0 & 3759.8 & 209.0 \\
 & decoder  & 13.3 & 26.6 & 405.0 & 23.7 \\
 & head  & 7.2 & 14.4 & 94.4 & 4.5 \\
\rowcolor{gray!20} \cellcolor{white} & \textbf{Total} & 1565.5\dplus{-31.9\%} & 3131.0\dplus{-31.9\%} & 4259.2\dplus{-22.2\%} & 237.2\dplus{-25.2\%}\\
\midrule
\midrule
\multirow{4}{*}{\tabincell{c}{Sparse4Dv2\\(Baseline)}} & Backbone  & 2280 & 4560.0 & 4977.5 & 278.8 \\
 & Decoder & 10.8 & 21.6 & 281.3 & 37.4 \\
 & Head & 1.7 & 3.4 & 7.5 & 5.8 \\
\rowcolor{gray!20} \cellcolor{white}  & \textbf{Total}  & 2292.5 & 4585.0 & 5266.3 & 322.0 \\
 \cmidrule{2-6}
\multirow{4}{*}{Ours-Faster} & Backbone  & 1545.0 & 3090.0 & 3,759.9 & 206.6 \\
 & Decoder & 10.8 & 21.6 & 281.3 & 32.3 \\
 & Head & 1.7 & 3.4 & 7.5 & 5.9 \\
\rowcolor{gray!20} \cellcolor{white}  & \textbf{Total}  & 1557.5\dplus{-32.1\%} & 3115.0\dplus{-32.1\%} & 4048.7\dplus{-23.1\%} & 244.8\dplus{-24.0\%} \\
\bottomrule 
\end{tabular}

\end{table}

We provide the MACs and FLOPs, as shown in the Tab.~\ref{tab:profile}. Our method significantly reduces the MACs and FLOPs by up to 32\% compared with the baseline methods, \ie, StreamPETR~\cite{wang2023streampetr} and Sparse4Dv2~\cite{lin2023sparse4dv2}. The results prove the computational efficiency of our method.

Besides, we conduct a profiling analysis of methods using the ViT-L backbone (also shown in the Tab.~\ref{tab:profile}). It shows that the ViT backbone is the bottleneck of the computational efficiency, which consumes nearly 90\% GFLOPs, GPU memory, and inference time of the whole detector. Our method reduces the resource consumption of the backbone by 30\% and significantly improves the efficiency of the whole detector.

\begin{figure}[h]
\begin{center}
\includegraphics[width=\linewidth]{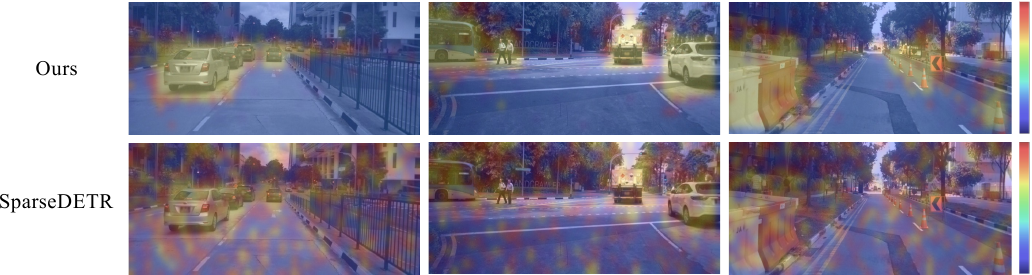}
\end{center}
\caption{The visualization of importance score from SparseDETR~\cite{roh2021sparsedetr} and our method. It is better viewed in color and zoomed in.}
\label{fig:supp_score_vis}
\end{figure}

\section{More Qualitative Results}
\label{sec: supp_vis}
\begin{figure}[t]
\begin{center}
\includegraphics[width=\linewidth]{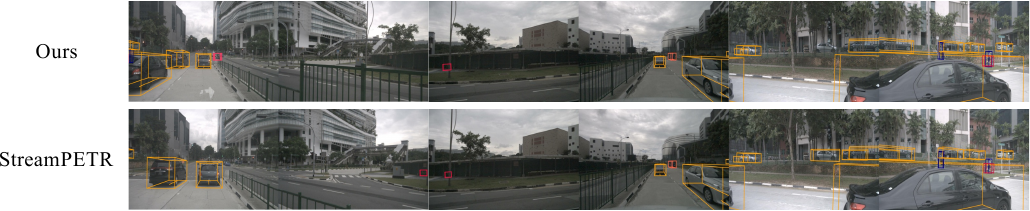}
\end{center}
\caption{The visualization of predictions from StreamPETR~\cite{wang2023streampetr} and our method. It is better viewed in color and zoomed in.}
\label{fig:supp_pred_vis}

\begin{center}
\includegraphics[width=\linewidth]{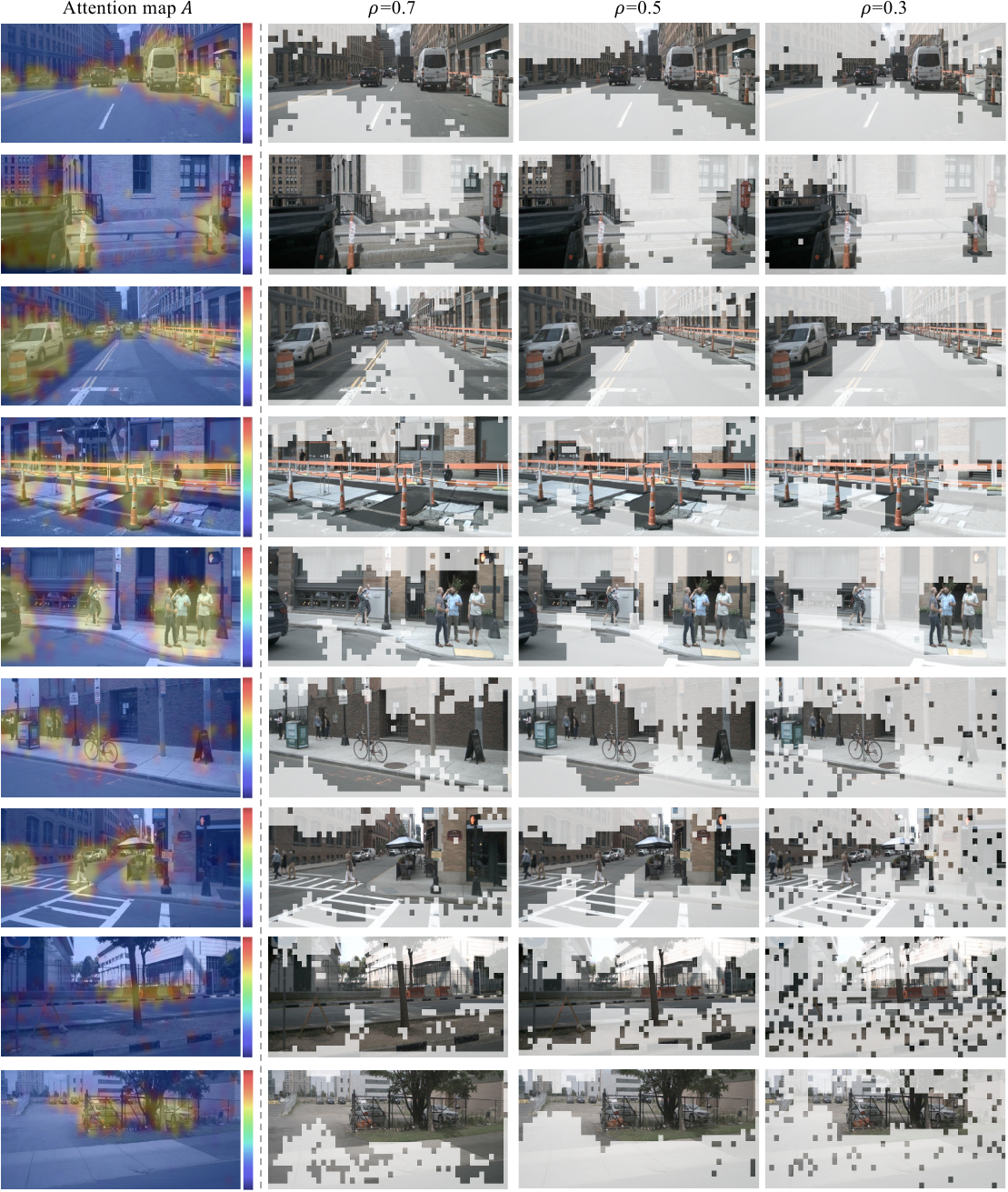}
\end{center}

\caption{The visualization of our method in various scenes (better viewed in color). We visualize the attention map in importance score calculation on the left and the salient/redundant tokens after the top-k selection on the right. Redundant tokens are illustrated as translucent.}
\label{fig: supp_vis}
\end{figure}

To better understand the superiority of our MQTS, we provide qualitative comparison results in Fig.~\ref{fig:supp_score_vis}, which clearly shows that MQTS can focus more on the foreground objects while SparseDETR~\cite{roh2021sparsedetr} fails to do that. We also visualize the predictions from our method and StreamPETR in Fig.~\ref{fig:supp_pred_vis}, proving that our method can significantly improve efficiency with nearly the same performance.

Additionally, we provide more visualization results for qualitative analysis, shown in Fig.~\ref{fig: supp_vis}. These results further prove our claim, \ie, the attention map models the correlations between image tokens and history queries, and thus can represent the foreground information density of each token since history queries contain foreground priors. With the 3D foreground object-aware attention map, the whole model can be more concentrated on foreground tokens when keeping ratios $\rho$ getting lower, improving the efficiency.

\clearpage
\clearpage

\end{document}